# Training a Functional Link Neural Network Using an Artificial Bee Colony for Solving a Classification Problems

Yana Mazwin Mohmad Hassim and Rozaida Ghazali

**Abstract**—Artificial Neural Networks have emerged as an important tool for classification and have been widely used to classify a non-linear separable pattern. The most popular artificial neural networks model is a Multilayer Perceptron (MLP) as is able to perform classification task with significant success. However due to the complexity of MLP structure and also problems such as local minima trapping, over fitting and weight interference have made neural network training difficult. Thus, the easy way to avoid these problems is to remove the hidden layers. This paper presents the ability of Functional Link Neural Network (FLNN) to overcome the complexity structure of MLP by using single layer architecture and propose an Artificial Bee Colony (ABC) optimization for training the FLNN. The proposed technique is expected to provide better learning scheme for a classifier in order to get more accurate classification result.

——————————— ◆ ———————————

## 1 INTRODUCTION

ARTIFICIAL Neural Networks have been known to be successfully applied to a variety of real world classification tasks especially in industry, business and science [1, 2]. Artificial Neural Networks (ANNs) especially the Multilayer Perceptron (MLP) are capable of generating complex mapping between the input and the output space in performing arbitrarily complex nonlinear decision boundaries. For a classification task, ANNs needs to be "trained" for the network to be able to produce the desired input-output mapping. In training phase, a set of example data are presented to the network and the connection weights of the network are adjusted by using a learning algorithm. The purpose of the weights adjustment is to enable the network to "learn" so that network would adapt to the given training data.

The most common architecture of ANNs is the multi-layer feedforward network (MLP). MLP utilize a supervised learning technique called Backpropagation for training the network. However, due to its multi-layered structure, the training speeds are typically much slower as compared to other single layer feedforward networks [3]. Problems such as local minima trapping, overfitting and weight interference also make the network training in MLP become challenging [4]. Hence, Pao [5] has introduce an alternative approach named Functional Link Neural Network (FLNN) in avoiding these problems. This approach removes the hidden layer from the ANN architecture to help in reducing the neural architectural complexity and provides them with an enhancement representation of input nodes for the network to be able to perfom a non-linear separable classification task[5, 6].

In this paper, we describe an overview of FLNN and the proposed Artificial Bee Colony (ABC) as learning algorithm in order to achieve better classification abilility. The rest of this paper is organized as follows: A background and related work regarding to the Multilayer Perceptron, FLNN and Population-based optimization technique are given in section 2. The proposed Population-based optimization for training the FLNN is detailed in section 3. The simulation result of the proposed learning scheme is presented in section 4. Finally, the paper is concluded in section 5.

## 2 RELATED WORK
### 2.1 Artificial Neural Network

Artificial Neural Networks (ANNs) are information processing model inspired by the way of human brain processes information. ANNs required knowledge through a learning process while the interneuron connection strength known as synaptic weights are used to store knowledge [7]. Therefore with these abilities, Neural Networks provides a suitable solution for pattern recognition or data classification problems.

One of the best known types of Neural Networks is the Multilayer Perceptron (MLP). It has one or more hidden layers in between the input and the output layer. Figure 1 illustrates the layout of MLP with single hidden layer. The function of hidden neurons is to provide the ANNs with the ability to handle non-linear input-output mapping. By adding one more hidden layer, the network is able to extract higher order statistics, which is particularly valuable when the size of the input layer is large [7].

————————————————

- *Yana Mazwin Mohmad Hassim is with the Universiti Tun Hussein Onn Malaysia, 86400 Parit Raja, Batu Pahat, Johor, Malaysia.*
- *Rozaida Ghazali is with the Universiti Tun Hussein Onn Malaysia, 86400 Parit Raja, Batu Pahat, Johor, Malaysia.*

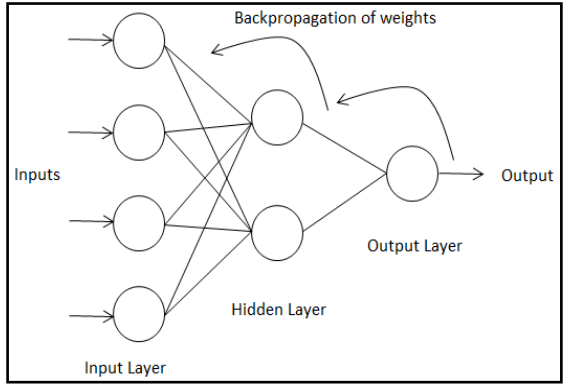

Fig. 1. The structure of single hidden layer MLP with Backpropagation algorithm

The MLP networks are trained by adjusting the weight of connection between neurons in each layer. For training the network, MLP utilize a supervised learning technique called Backpropagation, in which the network is provided with examples of the inputs and desired outputs to be computed, and then the error (difference between actual and expected results) will be calculated. Figure 1, depicted an example of MLP with Backpropagation. With this architecture, the neurons are organized in layers and their signals are sent "forward" while the calculated errors are then propagated backwards. The idea of the backpropagation algorithm is to reduced error, until the networks learned the training data. The training began with random weights, and the goal is to adjust them until the minimal error is achieved.

## 2.2 Higher Order Neural Network

Higher order Neural Networks (HONNs) is a different type of neural network with the presence of expanded input space in it single layer feed-forward architecture. HONNs contain summing unit and product units that multiply their inputs. These high order terms or product units can increase the information capacity for the input features and provides nonlinear decision boundaries to give a better classification capability than the linear neuron [8]. A major advantage of HONNs is that only one layer of trainable weight is needed to achieve nonlinear separable, unlike the typical MLP or feed-forward neural network [9].

Although most neural networks models share a common goal in performing functional mapping, different network architecture may vary significantly in their ability in handle different types of problems. For some tasks, higher order architecture of some of the inputs or activations may be appropriate to help form good representation for solving the problems. HONNs are needed because ordinary feed-forward network like MLP cannot avoid the problem of slow learning, especially when involving highly complex nonlinear problems [10].

### 2.2.1 Functional Link Neural Network

Functional Link Neural Network (FLNN) is a class of Higher Order Neural Networks (HONNs) that utilize higher combination of its inputs [5, 6]. It was created by Pao [6] and has been successfully used in many applications such as system identification [11-16], channel equalization [3], classification [17-20], pattern recognition [21, 22] and prediction [23, 24]. In this paper, we would discuss on the FLNN for the classification task. FLNN is much more modest than MLP since it has a single-layer network compared to the MLP but still is able to handle a non-linear separable classification task. The FLNN architecture is basically a flat network without any hidden layer which has make the learning algorithm used in the network less complicated [9]. In FLNN, the input vector is extended with a suitably enhanced representation of the input nodes, thereby artificially increasing the dimension of the input space [5, 6].

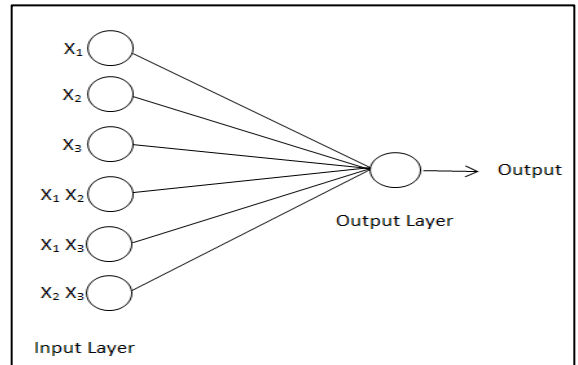

Fig. 2. The 2nd order FLNN structure with 3 inputs

Our focused on this work is on Functional link neural networks with generic basis architecture. This architecture uses a tensor representation. In tensor model the input features of x for example $x_i$, can be enhance into $x_i x_j$, $x_i x_j x_k$ and more with condition of $i \leq j \leq k$. The enhanced input features from the input layer are fed into the network and the output node $s$, in the output layer calculates the weighted sum of inputs and passes it through activation function to produce network output, $y$. Pao [6], Patra [12], Namatamee [25] has demonstrated that this architecture is very effective for classification task. Figure 2 depicts the functional link neural network structure up to second order with 3 inputs. The first order consist of the 3 inputs $x_1$, $x_2$ and $x_3$, while the second order of the network is the extended input based on the product unit $x_1 x_2$, $x_1 x_3$, and $x_2 x_3$. The learning part of this architecture on the other hand, consists of a standard Backpropagation as the training algorithm.

In most previous researches, the learning algorithm used for training the FLNN is the Backpropagation (BP) [4]. Table I below summarizes recent studies on FLNN for classification.

TABLE 1
previous research on FLNN Training[4]

| Reference | Method | Learning algorithm |
|---|---|---|
| [26] | ClFLNN | Pseudoinverse |
| [27] | EFLN | BP-learning |
| [28] | ClaFLNN | Genetic Algorithm with BP-learning |
| [18] | GFLNN | Adaptive learning |
| [9] | FLANN | BP-learning |
| [29] | ClasFLNN | BP-learning |
| [24] | FLNN | BP-learning |

Even though BP is the mostly used algorithm in training the FLNN, the algorithm however, has several limitations. First, it is easily get trapped in local minima especially for those non-linearly separable classification problems. Second, the convergence speed of the BP learning is too slow even if the learning goal can be achieved. Third, the convergence behavior of the BP algorithm depends on the choices of initial values of the network connection weights as well as the parameters in the algorithm such as the learning rate and momentum [4]. For these reasons, a further investigation to improve a learning algorithm used in tuning the learnable weights in FLNN is desired.

### 2.3 Artificial Bee Colony

The Artificial Bees Algorithm is an optimization tool, which provides a population-based search procedure [30]. The ABC algorithm simulates the intelligent foraging behavior of a honey bee swarm for solving multidimensional and multimodal optimization problem [31]. In population-based search procedure, each individual population called foods positions are modified by the artificial bees while the bee's aim is to discover the places of food sources with high nectar amount and finally the one with the highest nectar.

In this model, the colony of artificial bees consists of three groups: which are employed bees, onlookers and scouts [32]. For each food source there is only one artificial employed bee. The number of employed bees in the colony is equal to the number of food sources around the hive. Employed bees go to their food source and come back to hive and with three information regarding the food source: 1) the direction 2) its distance from the hive and 3) the fitness and then perform waggle dance to let the colony evaluate the information. Onlookers watch the dances of employed bees and choose food sources depending on dances. After waggle dancing on the dance floor, the dancer goes back to the food source with follower bees that were waiting inside the hive. This foraging process is called local search method as the method of choosing the food source is depend on the on the experience of the employed bees and their nest mates [31]. The employed bee whose food source has been abandoned becomes a scout and starts to search for finding a new food source randomly without using experience. If the nectar amount of a new source is higher than that of the previous one in their memory, they memorize the new position and forget the previous one [31]. This exploration managed by scout bees is called global search methods.

Several studies done by [31-33] has described that the Artificial Bee Colony algorithm is very simple, flexible and robust as compared to the existing population-based optimization algorithms: Genetic Algorithm (GA), Differential Evolution (DE) and Particle Swarm Optimization (PSO) in solving numerical optimization problem. As in classification task in data mining, ABC algorithm also provide a good performance in gathering data into classes [34]. Hence motivated by these studies, the ABC algorithm is utilized in this work as an optimization tool to optimize FLNN learning for solving classification problem.

## 3 PROPOSED TRAINING SCHEME

Inspired by the robustness and flexibility offered by the population-based optimization algorithm, we proposed the implementation of the ABC algorithm as the learning scheme to overcome the disadvantages caused by backpropagation in the FLNN training. The proposed flowchart is presented in figure 4. In the initial process, the FLNN architecture (weight and bias) is transformed into objective function along with the training dataset. This objective function will then be fed to the ABC algorithm in order to search for the optimal weight parameters. The weight changes are then tuned by the ABC algorithm based on the error calculation (difference between actual and expected results).

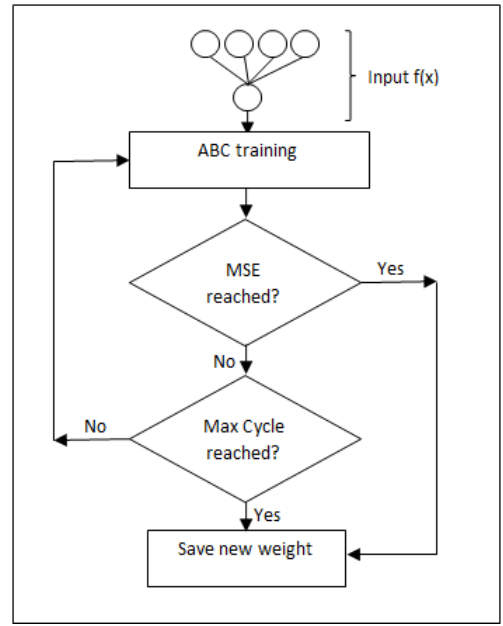

Fig. 4. The proposed Training scheme for FLNN-ABC

Based on the ABC algorithm, each bee represents the solutions with a particular set of weight vector. The ABC algorithm [35] used for training the FLNN is summarized as follow:
1. Cycle 0:
2. Initialize a population of scout bee with random solution $x_i$, $i = 1,2, .. SN$
3. evaluate fitness of the population
    a. initialize weight and bias for the FLNN
4. cycle 1: while Maximum cycle not reached,
   repeat step 5- step 11
5. form new population $v_i$ for the employed bees using:

$$v_{ij} = x_{ij} + \varphi_{ij}(x_{ij} - x_{kj})$$

where $k$ is a solution in the neighbourhood of $i$, $\Phi$ is a random number in the range [-1,1] and evaluate them.
    a. Form new weight and bias for FLNN
6. Calculate fitness function on new population

$$fit_i = \begin{cases} \frac{1}{1+f_i}, & if\ f_i \geq 0 \\ 1 + abs(f_i), & if\ f_i < 0 \end{cases}$$

7. Apply the greedy selection process between $x_{ij}$ and $v_{ij}$
8. Calculate the probability values $p_i$ for the solutions $x_i$ using:

$$p_i = \frac{fit_i}{\sum_{n=1}^{SN} fit_n}$$

9. If the onlookers are distributed
    a. Go to step 11
10. else
    a. Repeat step 5 until step 8 for onlookers.
11. Apply the greedy selection process for onlookers, $v_i$
12. Determine the abandoned solution for the scout, if exists, and replace it with a new randomly produced solution $x_i$ using:

$$x_i^j = x_{min}^j + rand(0,1)(x_{max}^j - x_{min}^j)$$

13. Memorize the best solution
    a. Update new weight and bias for the FLNN
14. cycle=cycle+1
15. Stop when cycle = Maximum cycle
16. Save updated FLNN new weight and bias.

## 4 SIMULATION RESULT

In order to evaluate the performance of the proposed learning scheme FLNN-ABC for classification problem, simulation experiments were carried out on a 2.30 GHz Core i5-2410M Intel CPU with 8.0 GB RAM in a 64-bit Operating System. The comparison of standard BP training and ABC algorithms is discussed based on the simulation results implemented in Matlab 2010b. In this work we considered 4 benchmark classification problems, Breast Cancer Wisconsin, PIMA Indian Diabetes and BUPA Liver Disorder.

During the experiment, simulations were performed on the training of MLP architecture with Backpropagation algorithm (MLP-BP), second order FLNN architecture with Backpropagation algorithm (2ndFLNN-BP) and second order FLNN architecture with ABC algorithm (2ndFLNN-ABC) as their learning scheme. The best training accuracy for every benchmark problems is taken out from these simulations. The Learning rate and momentum used for both MLP-BP and 2ndFLNN-BP were 0.3 and 0.7 with the maximum of 1000 epoch and minimum error=0.001 as for the stopping criteria. Parameters setup for the 2ndFLNN-ABC however, only involved the setting up of stopping criteria of maximum 100 cycles and minimum error=0.001. The activation function used for the network output for both MLP and 2ndFLNN is Tangent Hyperbolic sigmoid function. Table 2 below summarized the parameters considered in this simulation.

TABLE 2
Parameters considered MLP-BP, 2nd order FLNN-BP and 2nd order FLNN-ABC simulation

| parameters | MLP-BP | 2ndFLNN-BP | 2ndFLNN-ABC |
|---|---|---|---|
| Learning rate | 0.3 | 0.3 | - |
| Momentum | 0.7 | 0.7 | - |
| Maximum epoch | 1000 | 1000 | - |
| Minimum error | 0.001 | 0.001 | 0.001 |
| Maximum cycle (MCN) | - | - | 100 |
| Parameters range | [-1,1] | [-1,1] | [-10,10] |

The reason of conduction a simulation on MLP and FLNN architectures is to provide a comparison on neural complexity between standard Artificial Neural Network and Higher Order Neural Network. In this simulation we used a single hidden layer of MLP with the numbers of hidden nodes equal to it input features while for the FLNN architecture we used a second order input enhancement (2ndFLNN) to provide a nonlinear input-output mapping capability. The neural complexity refers to the numbers of trainable parameters (weight and bias) needed to implement good approximation in each neural network architecture. The less numbers of parameters indicate that the network required less computational load as there are small numbers of weight and bias to be updated at every epoch or cycle and thus led to a faster convergence. Table 3 below shows the comparison of network complexity for each benchmark problems.

Ten trials were performed on each simulation of the MLP-BP, 2ndFLNN-BP and 2ndFLNN-ABC with the best training accuracy result is taken out from these 10 trials. In order to generate the Training and Test sets, each datasets were randomly divided into two equal sets (1st-Fold and 2nd-Fold). Each of these two sets was alternatively used either as a Training set or as a Test set. The average values of each datasets result were then used for comparison. Table 4 below, presents the simulation result of MLP-BP, 2ndFLNN-BP and 2ndFLNN-ABC architectures. It is shown that, the implementation of ABC as the training scheme for FLNN-ABC provides better accuracy on Test set compared to FLNN-BP and MLP-BP.

TABLE 3
Comparison on complexity of the architecture between MLP and FLNN for each benchmark datasets

| Datasets | Network type | Network structure | No of parameters/dimensions (D) |
|---|---|---|---|
| Cancer | MLP | 9-9-1 | 100 |
|  | 2nd FLNN | 45-1 | 46 |
| PIMA | MLP | 8-8-1 | 83 |
|  | 2nd FLNN | 36-1 | 37 |
| BUPA | MLP | 6-6-1 | 49 |
|  | 2nd FLNN | 21-1 | 22 |

TABLE 4
Result obtained from MLP-BP, 2nd FLNN-BP and 2nd FLNN-ABC architectures

| Datasets | Average Values | MLP-BP | 2nd FLNN-BP | 2nd FLNN-ABC |
|---|---|---|---|---|
| Cancer | MSE on Training | 0.19186 | 0.24311 | 0.071645 |
|  | Training Accuracy (%) | 90.40625 | 87.84335 | 93.9638 |
|  | MSE on Test | 0.21442 | 0.241233 | 0.180275 |
|  | Test Accuracy (%) | 89.26655 | 75.5669 | 90.5899 |
| PIMA | MSE on Training | 0.697915 | 0.539665 | 0.41128 |
|  | Training Accuracy (%) | 65.10415 | 72.7838 | 74.33805 |
|  | MSE on Test | 0.6979165 | 0.5965245 | 0.580515 |
|  | Test Accuracy (%) | 65.10415 | 39.5953 | 69.8524 |
| BUPA | MSE on Training | 0.84087 | 0.853635 | 0.51649 |
|  | Training Accuracy (%) | 57.9564 | 57.06855 | 54.99575 |
|  | MSE on Test | 0.840872 | 1.075618 | 0.726065 |
|  | Test Accuracy (%) | 57.9564 | 17.08535 | 59.1484 |

## 5 CONCLUSION

In this work, we evaluated the FLNN-ABC model for the task of pattern classification of 2-class classification problems. The experiment has demonstrated that FLNN-ABC performs the classification task quite well. For the case of Cancer, PIMA and BUPA the simulation result shows that the proposed ABC algorithm can successfully train the FLNN for solving classification problems with better accuracy rate on unseen data. The importance of this research work is to provide an alternative training scheme for training the Functional link Neural Network instead of standard BP learning algorithm.

**Yana Mazwin Mohmad Hassim** is a Ph.D student at Universiti Tun Hussein Onn Malaysia (UTHM). Her current research focuses on the optimization of Higher Order Neural Networks using Swarm Intelligence Algorithms. She received her Master's Degree in Computer Science from University of Malaya in 2005 and Bachelor's Degree in Information Technology from National University of Malaysia in 2001.

**Rozaida Ghazali** is a Deputy Dean (Research and Development) at the Faculty of Information Technology and Multimedia, Universiti Tun Hussein Onn Malaysia (UTHM). She graduated with Ph.D. degree from the School of Computing and Mathematical Sciences at Liverpool John Moores University, United Kingdom in 2007, on the topic of Higher Order Neural Networks for Financial Time series Prediction. Earlier, in 2003 she completed her M.Sc. degree in Computer Science from Universiti Teknologi Malaysia (UTM). She received her B.Sc. (Hons) degree in Computer Science (Information System) from the Universiti Sains Malaysia (USM) in 1997. In 2001, Rozaida joined the academic staff in UTHM. Her research area includes neural networks, data mining, financial time series prediction, data analysis, physical time series forecasting, and fuzzy logic.